\documentclass[]{interact}

\usepackage{epstopdf}% To incorporate .eps illustrations using PDFLaTeX, etc.
\usepackage[caption=false]{subfig}% Support for small, `sub' figures and tables
\usepackage[acronym]{glossaries}
\usepackage[natbibapa,nodoi]{apacite}
\theoremstyle{plain}% Theorem-like structures provided by amsthm.sty

\theoremstyle{definition}

\theoremstyle{remark}

\makeglossaries

\newacronym{hei}{HEI}{higher education institution}
\newacronym{nlp}{NLP}{Natural Language Processing}
\newacronym{bip}{BIP}{blended intensive program}
\newacronym{procot}{ProCoT}{Probing Chain-of-Thought}
\newacronym{llm}{LLM}{Large Language Model}
\newacronym{ai}{AI}{Artificial Intelligence}
\newacronym{ml}{ML}{machine learning}
\newacronym{ilos}{ILOs}{intended learning outcomes}
\newacronym{rag}{RAG}{Retrieval Augmented Generation}

\begin{document}

\articletype{ARTICLE}% Specify the article type or omit as appropriate

\title{\acrshort{procot}: Stimulating Critical Thinking and Writing of Students through Engagement with Large Language Models (\acrshort{llm}s)}

\author{
\name{Tosin Adewumi\textsuperscript{*}\thanks{https://orcid.org/0000-0002-5582-2031}, Lama Alkhaled, Claudia Buck, Sergio Serrano Hernández, \\ Saga Brilioth, Mkpe O. Kekung, Yelvin Ragimov, and Elisa Barney. \textsuperscript{}}
\affil{\textsuperscript{}Luleå University of Technology (LTU), Sweden; \\ \textsuperscript{*}Corresponding author (tosin.adewumi@ltu.se)}
}

\maketitle

\begin{abstract}
We introduce a novel writing method called \textit{\acrfull{procot}}, which potentially prevents students from cheating using a \acrfull{llm}, such as ChatGPT, while enhancing their active learning.
%through such models.
\acrshort{llm}s have disrupted education and many other fields.
For fear of students cheating, many have resorted to banning their use.
%, as their outputs can be human-like and hard to detect in some cases.
These \acrshort{llm}s are also known for hallucinations.
%(i.e. fake “facts”).
We conduct studies with \acrshort{procot} in two different courses with
%a combined total of 
65 students.
%\acrshort{procot} is entrenched in the \textit{self-regulation} method.
The students in each course were asked to prompt an \acrshort{llm} of their choice with one question from a set of four and required to affirm or refute statements in the \acrshort{llm} output by using peer-reviewed references.
The results show two things: (1) \acrshort{procot} stimulates creative/critical thinking and writing of students through engagement with \acrshort{llm}s when we compare the \acrshort{llm}-only output to \acrshort{procot} output and (2) \acrshort{procot} can prevent cheating because of clear limitations in existing \acrshort{llm}s, particularly ChatGPT, when we compare students' \acrshort{procot} output to \acrshort{llm} \acrshort{procot} output. 
We also discover that most students prefer to give answers in fewer words than \acrshort{llm}s, which are typically verbose.
The average word counts for students in the first course, ChatGPT (v3.5), and Phind (v8) are 208, 391 and 383, respectively.

\end{abstract}


% Back then had an essay to compare what a short summary of a larger book says about the original book - Cliffs Notes

\section{Introduction}
A chain of thought is a sequence of intermediate reasoning steps \citep{wei2022chain}.
Chain-of-thought (CoT) prompting allows a sufficiently \acrfull{llm} to solve a problem as a clear sequence of steps before a final answer.
An \acrshort{llm}, which is typically a deep neural probabilistic model \citep{brown2020language}, such as ChatGPT\footnote{openai.com}, has the ability to generate convincing outputs when given inputs, called prompts.
The use of such \acrfull{ai} tools, particularly ChatGPT, has gained immense popularity in many fields in the last one year  \citep{Allam2023}.
They have also gained notoriety for suspected cheating, particularly in the Education field \citep{cotton2023chatting,ventayen2023chatgpt}.
This is besides the \textit{hallucinations} (or ungrounded addition) problem with \acrshort{llm}s \citep{rawte2023survey,info13060298}, which occurs partly because they are typically trained on static data, with no updates through time.
Humans, on the other hand, can source and use real-world context to verify the veracity of claims \citep{maslej2023artificial}.

Topic-based writing has always been a common way of assessing students \citep{todd2004measuring}.
However, the capability of \acrshort{llm}s to completely or partially generate content as desired by a user appears to be putting a threat to students writing own their own.
The research question we address in this work is: `\textbf{Is there a way to both prevent cheating in writing by students through \acrshort{llm}s and also engage them in the use of such \acrshort{ai} tools?}'
The generative power of these \acrshort{llm}s makes it imperative to ask similar questions regarding coding, mathematics and other similar tasks.

We introduce \textit{\acrfull{procot}}, a method aimed at stimulating creative/critical thinking and writing of students through engagement with \acrshort{llm}s, given that these models are here to stay.
We demonstrate this method in two University courses (an Erasmus+ \acrfull{bip} \citep{ecBIP2022} and a full semester course) for two sets of students.
The students were free to use the \acrshort{llm} of their choice and 95\% used ChatGPT\footnote{openai.com} (v3.5) while 5\% used Phind\footnote{phind.com} (v8) in the \acrshort{bip}.
Similarly, most of the students in the second course used ChatGPT (96\%) while BingAI\footnote{bing.com/search?q=Bing\%20AI\&showconv=1\&form=MA13FV (limited to 2,000 characters of input)} was 4\%.
We believe \acrshort{procot} will empower educators to no longer fear \acrshort{llm}s but see them as technological tools and students will be better critical thinkers.
Also, it may be applied to any field that uses writing.

%Shortcomings exist in existing fact-checking datasets.
%Several of them contain claims that lack sufficient evidence or
%counter-evidence that exist in a trusted knowledge base.

%In order to validate whether or not certain assertions in the output of an \acrshort{llm} is 'true', we can either check for it in the supplied prompt ('narrow grounding') or use general knowledge ('broad grounding').

% \cite{rezaei2010reliability}

\subsection{Our Contributions}
\begin{enumerate}
    \item \acrshort{procot} prevents cheating through \acrshort{llm}s by students.
    \item \acrshort{procot} stimulates creative/critical thinking and writing of students through engagement with \acrshort{llm}s
    \item \acrshort{procot} will provide valuable data to further train \acrshort{llm}s to be better models without privacy issues.
\end{enumerate}

The rest of this paper is organized as follows.
Section \ref{back} discusses the background and literature review, including \acrshort{llm}s.
Section \ref{method} describes in detail the methods we follow, including the two cases.
Section \ref{result} discusses the outcome in results that also present statistical analyses.
Section \ref{conc} briefly gives concluding remarks.

\section{Background and Literature Review}
\label{back}

In essay writing assessments, students are asked to demonstrate a point of view \citep{todd2004measuring} and this may be supported by evidence (e.g. through peer-reviewed references).
This allows students to show what they have learnt and build their writing skills.

\subsection{Essay Writing and Evaluation}
The pedagogy of essay writing evaluation requires a comprehensive approach that focuses on student learning and development.
Central to this is the use of rubrics, which are detailed guides laying out the expected criteria for an essay \citep{lipnevich2023unraveling}.
They may cover aspects like the strength of an argument, the use of evidence (or references), and the organization of ideas \citep{todd2004measuring}.
%, and the quality of writing.
%This process might start with formative assessment, which emphasizes continuous feedback throughout the essay writing process. Unlike summative assessments that provide a final grade, formative assessments are ongoing and interactive, allowing students to understand, reflect upon, and integrate feedback into their learning journey.
A formative approach to evaluation, based on quality feedback, helps students identify their strengths and weaknesses and fosters a growth mindset
%, encouraging them to view writing as a skill that can be developed over time 
\citep{spector2016technology,porter2022collaborative}.
These methods also provide transparency and ensure alignment with \acrfull{ilos} \citep{le2023impact, ramesh2022automated}.
%The transparency provided by rubrics demystifies the grading process, offering students a clear understanding of what is expected and how their work will be evaluated. This clarity not only aids in self-assessment but also ensures that the feedback provided is structured and .

Peer review and self-assessment are also important in pedagogy.
By engaging in peer review, students develop critical thinking skills
%, as they learn to evaluate others' work constructively 
\citep{cheong2023self} 
%€This process not only deepens their understanding of good writing practices but also promotes a collaborative learning environment \citep{chambers2023increasing}.
while self-assessment encourages students to reflect on their own writing, fostering a sense of ownership and responsibility \citep{culver2023learning}.
%Quality feedback is another cornerstone of this approach. Effective feedback is not just about pointing out flaws; it's about providing constructive, specific, and actionable suggestions that guide students toward improvement. This type of feedback helps students understand the nuances of good writing and critical thinking, essential skills in their academic and professional lives \citep{noroozi2023design}.
Encouraging reflection is vital in the pedagogical evaluation of any writing.
%Reflective practice involves students contemplating their writing process and the feedback received, leading to self-regulated learning.
It helps students to be more independent and confident writers
%, capable of critical analysis and self-improvement 
\citep{pang2022teachers}.
The pedagogical evaluation of writing is a comprehensive yet subjective process \citep{todd2004measuring}.
%that places a strong emphasis on enhancing student learning and growth. Adopting this method 
It is 
%transforms essay assessment into 
a crucial component of the educational journey that plays a significant role in the overall development of students, both as critical thinkers and skilled writers.

\subsection{Large Language Models (\acrshort{llm}s)}

%A pivotal technological advancement which has been universally implemented in recent years,
An \acrshort{llm}
%can be defined as a carefully engineered AI model with a clear 
aims to mimic human language patterns, responses, and general linguistic structures after training.
It is a deep probabilistic model that is trained on big data to generate probabilities over a set of tokens (or words) for predicting the next token in a sequence and there are many types with different sizes \citep{brown2020language, touvron2023llama,10191208}.
%present in the language in which it is trained on. \acrshort{llm}s constitute a part of a larger group called transformers – a type of neural network (or a deep learning model simulating the learning pathway of the human brain) which develops an understanding of context and significance by analyzing sequential data relationships, such as that between words in a sentence. 
%According to OpenAI’s own transformer Large Language Model, ChatGPT, an LLM 
%\begin{quote}
 %   “refers to a type of \acrfull{ai} model designed to understand and generate human-like language on a large scale. These models are built using deep learning techniques, specifically using architectures like transformers. The term "large" emphasizes the extensive size of the model, both in terms of the number of parameters and the data it has been trained on.” - ChatGPT\footnote{davicci on December 10, 2023}
%\end{quote}
%In other words, OpenAI’s ChatGPT adheres to the above understanding of the premise behind \acrshort{llm}s, all the while being  a clear and 
%As mentioned earlier, Transformers, and so Large Language Models, are included in the larger concept of Neural Network Models: an advanced training model comprised of multiple complementary components, which together perform advanced tasks such as pattern recognition, data-based prediction and speech recognition, just as with the human brain. Its name derives from its purpose of resembling the knowledge processing and learning curve found in the brain’s neural networks.
These models are usually based on the Transformer, which is a deep neural network architecture \citep{vaswani2017attention}.
They are trained to perform \acrfull{nlp} tasks, like reading comprehension, summarization, and question answering \citep{radford2019language,touvron2023llama}.
%are, in essence, revealing the closest current opportunity of knowledge processing similarly to that of a human
%; in other words, Neural Networks give Artificial Intelligence a brain.
%Neural Networks can subsequently be placed in a larger category called Deep Learning Architectures. 
%Similar to the traditional understanding of the word, architecture within the context of artificial intelligence refers to the essential structure, or build-up, of the final functional creation – in this case the implemented deep learning framework. %Moreover, just as with infrastructural design, the architecture types will diverge from one another depending on their specific intention and assigned task. Neural networks thus only form one example of an architecture type amongst many others, and there is an encapsulating relationship going from these deep learning architectures all the way into the narrowed-down world of large language models (see Figure 1).

By far, the most popular example of an \acrshort{llm} is OpenAI's ChatGPT\textsuperscript{2}, though there are several others, like Gemini\footnote{deepmind.google/technologies/gemini} by Google DeepMind and LLAMA-2 \citep{touvron2023llama} by Meta.
ChatGPT, now in vesion 4 \citep{Natalie2023}, is an advanced \acrshort{llm}.
%This transition signifies a notable advancement in the capabilities of the ChatGPT model, being more creative and collaborative than in the previous version.
%ChatGPT is introduced as a class of machine learning which uses the text-based input of users to 
It is the latest in the family of Generative Pre-trained Transformer (GPT) models, which began with GPT-1 while version 2 \citep{radford2019language} in 2019 is recognized as possibly the first \acrshort{llm} because of its size of 1.5B parameters \citep{maslej2023artificial}.
%ChatGPT is a type of artificial intelligence.
%GPT-1 pioneered pre-training on various text on websites, learning patterns, books, and articles including up to 117 million parameters.
%In contrast, the new version, ChatGPT 4, includes up to 100 trillion parameters, showing immense improvement \citep{Ray2023}.
ChatGPT (v3.5) is available for free to users through a web interface or for a fee after a certain period through an Application Programming Interface (API).

\acrshort{llm}s offer a myriad of benefits, including assistance in learning, improvement in language skills, access to extensive information, and the provision of customized learning experiences.
Particularly, the capabilities of ChatGPT to generate coherent and contextually relevant text based on prompts have made it popular.
Similarly to other \acrfull{ml} models \citep{yamamoto2018automated}, it may be used to assist in evaluating student essays and providing feedback.
%that is not only consistent but also reflective of a deep understanding of content and structure.
%These models facilitate an interactive and dynamic learning environment, providing instant feedback and a vast reservoir of information, which can be adapted to individual learning styles and needs.
The use of \acrshort{llm}s in education is not without challenges, however \citep{sullivan2023chatgpt,sok2023chatgpt,meyer2023chatgpt}.
Over-reliance on these models can potentially lead to diminished critical thinking abilities in students.
%This reliance risks students becoming dependent on AI for answers, bypassing the deep understanding of concepts.
%Furthermore, there is the issue of potential plagiarism.
%, where students might misuse these tools for assignment generation, raising concerns about academic integrity.
%lama 
%This feature is particularly remarkable as it includes the creation of various forms of content, ranging from essays and stories to code and poetry. Despite its extensive training, ChatGPT is designed with privacy and safety in mind, ensuring that it does not learn or retain personal information from individual interactions.
%This aspect underscores the ethical considerations inherent in the development of such advanced \acrshort{ai} technologies.
%The model's ability to generate coherent, contextually relevant responses allows 
%This capability is a boon for educators, as it offers an efficient and effective means to guide students in improving their writing skills.
%Additionally, the model's adherence to privacy and safety standards ensures that personal information is not learned or retained during interactions, a critical aspect in educational environments.
Another critical aspect to consider is the inherent biases in \acrshort{llm}s \citep{meyer2023chatgpt,info13060298,philosophies4030041}.
Since these models are trained on existing data, they may unintentionally reflect and reinforce any biases that exist in the data they were trained on, which could affect the quality and fairness of information provided to students. %That underscores the need for ongoing monitoring and refinement of these models to ensure a balanced and unbiased learning experience.
%In terms of ethical considerations, the deployment of \acrshort{llm}s in educational settings calls for careful contemplation.
%Educators and policymakers must address the ethical use of these models, especially concerning data privacy and the consequences of sharing sensitive information. 
Ensuring responsible AI use is paramount to maintaining a safe and productive learning environment \citep{vinuesa2020role}.
%It is crucial to approach the integration of \acrshort{llm}s with a balanced perspective. This involves empowering students to use these tools effectively, fostering critical thinking, creativity, and ethical understanding, while also addressing the challenges and limitations of \acrshort{llm}s. The role of educators in guiding students on the responsible use of these technologies is crucial for enriching the educational experience and ensuring the ethical development of \acrshort{ai} technology in the learning process.

%\subsection{What is ChatGPT?}

\subsection{\acrshort{llm} Fact-Checking Methods}

%With regards to the benefits and limitations to student usage of \acrshort{llm}s, these models as a whole exhibit notable proficiency in various natural language processing tasks, such as text completion, translation, summarization, question answering, and even creative writing.
%Nonetheless, one must consider the presence of limitations within this field as with any other, namely the presence of bias and ethical concerns within the potentials of \acrshort{llm}s.

In recent years, \acrshort{llm} fact-checking methods have emerged for combating misinformation.
\acrfull{rag} models \citep{lewis2020retrieval}, which combine information retrieval and generation, and special-purpose knowledge bases or datasets are useful in this regard.
%Leveraging the power of \acrshort{nlp} and advanced \acrshort{ml}techniques, \acrshort{llm} have proven to be promising when it comes to fact-checking in an automated way, as they streamline an evaluation process of the claims and statements presented in the text to check its accuracy.
%Large Language Models, such as GPT-3.5 [\cite{brown2020language}], have demonstrated a remarkable capacity to understand and generate human-like text,
%in fact, a large part of this very article has been generated using a Large Language Model and then checked by a human. In the realm of fact-checking, these models are trained on diverse datasets to develop the ability to discern between true and false information, giving it the ability to fact-check a text that has not been previously seen. 
%Below, we will present some of the techniques used by LLMs that help in the context of fact checking:
%So far we have described what fact-checking means when it comes to LLMs, we have presented the reader with a set of methods that is commonly used for fact-checking, and now we will introduce 
A few datasets that are used for fact-checking in \acrshort{llm}s are listed below.
The availability of these methods have not made \acrshort{llm}s error-free, indicating that more work still needs to be done by the research community.
%for the sake of exemplification, although more exist and at the time of reading this article there might be much more commonly used ones.

\begin{enumerate}
    \item FEVER \citep{Thorne18Fever} - Fact Extraction and VERification (FEVER) is a widely used dataset that consists of claims and evidence, allowing \acrshort{llm}s to learn to validate or debunk statements based on the supporting information.
    The statements from the dataset are from Wikipedia and claims are classified as \textit{Supported}, \textit{Refuted} or \textit{NotEnoughInfo}.
    
    \item ClaimBuster \citep{arslan2020benchmark} - ClaimBuster is a dataset of 23,533 statements that is designed to identify claims which are worth fact-checking.
    %and the people who have created it claim that ClaimBuster has been used to create the first ever end-to-end automated fact-checking system.
    %This dataset contributes to the development of LLMs capable of fact-checking within the sources of digital or traditional media. 
    The statements are extracted from U.S. general election presidential debates.
    %This dataset is publicly available to the research community. The dataset contains around 23000 sentences which are divided in
    There are three possible categories in the data: \textit{non-factual}, \textit{unimportant factual statement}, and \textit{check-worthy factual statement.}

    %\item Fact-checking organizations:   Researchers leverage Existing resources from fact checking websites that do not make datasets themselves but investigates claims, rumors, and urban legends. Two examples of organizations that are behind those investigations are \cite{snopes} and \cite{politifact}.
\end{enumerate}

\section{Methodology}
\label{method}
Probing the chain-of-thought (output) of an \acrshort{llm} involves students scrutinizing the output by using a reference-based platform to provide time-relevant fact-checked feedback on the output.
The method is entrenched in the `Self-Regulation' method \citep{zimmerman2013theories}, which is the self-directive process by which learners transform their mental abilities into task-related skills.
It involves the following components in an iterative process: \textit{forethought}, \textit{planning}, \textit{monitoring}, \textit{failure detection} and \textit{correction}.
It leads to faster independence and self-awareness of students.
ProCoT may even be applied to other outputs, besides those of \acrshort{llm}s.
%\subsection{\acrshort{procot}}

We compare the quality of students’ writing, as evaluated by a human subject expert and using statistical analysis, to establish the effectiveness of ProCoT.
We compare 1) \acrshort{llm}-only outputs, 2) students' \acrshort{procot} outputs, and 3) \acrshort{llm} \acrshort{procot} outputs.
\acrshort{llm}-only output is the result of asking the model any of the the original questions among the lists in the sub-sections below.
Meanwhile, a student's \acrshort{procot} output is the feedback provided (with peer-reviewed references) for the \acrshort{llm}-only output while an \acrshort{llm} \acrshort{procot} output is the \acrshort{llm} feedback to their original (\acrshort{llm}-only) output.
We give the same instruction that is given to the students to the \acrshort{llm} when generating the \acrshort{llm} \acrshort{procot} output.
This is to determine if the students can cheat by generating \acrshort{procot} feedback successfully with the \acrshort{llm} and pass them off as their own.
These evaluations are carried out in two cases involving two different courses and two different sets of students.

\subsection{Case 1}
The first case involves 26 students, where about 88\% are Masters students, 9\% Bachelor's, and 3\% PhD, at a \acrshort{bip}.
A \acrshort{bip} is a short programme that combines physical mobility with a virtual part.
The 6-day \acrshort{bip} was titled `DigiHealth‐AI: Practice, Research, Ethics, and Regulation' and organized
%by Deggendorf Institute of Technology (DIT, Germany) 
in collaboration with partners from four European countries.
%These include Luleå University of Technology (LTU, Sweden), AI Sweden, University of Agder (UiA, Norway),  Czech Technical University (CTU, Czechia), and Technical University of Catalonia (UPC, Spain).

The students were asked to pose one out of the four questions below to any \acrshort{llm}.
The questions were randomly selected from two blogs.\footnote{wtamu.edu/~cbaird/sq/category/health; magazine.utoronto.ca/research-ideas}
Thereafter, the \acrshort{procot} instruction was `\textit{Write 1 page to affirm or refute assertions/statements made by ChatGPT/LLM in the response by using references from peer-reviewed articles}'.
The students were under supervision for a 30-minute period while providing answers.

\begin{enumerate}
    \item Did cancer exist before man-made chemicals were around to create it?

    \item Who will benefit from AI in healthcare?

    \item How long do you have to exercise for it to count?

    \item How will we avoid machine bias?

\end{enumerate}

\subsection{Case 2}
This case had 39 students who were all Masters students.
The course is titled `Text Mining' and is a regular semester course. %at Luleå University of Technology (LTU).
Again, each student was asked to pose one out of the four questions below to any \acrshort{llm}.
The questions were randomly selected from a list of 50 exam questions, which are dedicated to the course.
Thereafter, the \acrshort{procot} instruction was `\textit{Write, at least, 300 words (excluding the reference section) to affirm or refute statements made by ChatGPT/LLM by using (at least 5) references from peer-reviewed articles or books}'.
The students were allowed to go with the task and submit within a week.

\begin{enumerate}
    \item What is a large language model (LLM)?

    \item What is the vanishing gradient problem and how is it tackled?

    \item Describe the two paradigms of using a pre-trained model in transfer learning: feature extraction and fine-tuning.

    \item Name two similarity measurement metrics for vectorized semantic representation. Which one may be preferred and why?

\end{enumerate}

\section{Results and Discussion}
\label{result}

\subsection{Case 1}
Two out of twenty-six submissions from the students were rejected for lack of compliance to the \acrshort{procot} instructions.
Results from the valid students' \acrshort{procot} answers show that \textbf{cheating can be prevented} by framing questions as we have done with \acrshort{procot}.
Comparing the \acrshort{llm} \acrshort{procot} feedback on its answers to the original questions\footnote{on 12 December, 2023}, ChatGPT expressly says `\textit{I cannot provide direct references to peer-reviewed articles}' on the one hand while saying `numerous studies in peer-reviewed literature support' its answer.
In one instance, though, it referenced a \textit{doi}, which directed to a paper that was less than suitable.
Phind, meanwhile, appears to be better at providing peer-reviewed references, though some references are blog posts.
However, it appears to lift the same content supplied to it as if regurgitating (in what may be considered plagiarism) more than paraphrasing the content or generating original response for engaging communication.
Therefore, the quality of students' \acrshort{procot} answers (on the basis of grounding by references) is better than those by ChatGPT, though ChatGPT usually gives more comprehensive answers that are typically verbose but not necessarily always factual.
They are also better than those by Phind on the basis of originality, though Phind may also give more verbose answers.

\begin{figure}
\centering
\subfloat[Box and Whisker plot of \acrshort{procot} number of words]{%
\resizebox*{5cm}{!}{\includegraphics{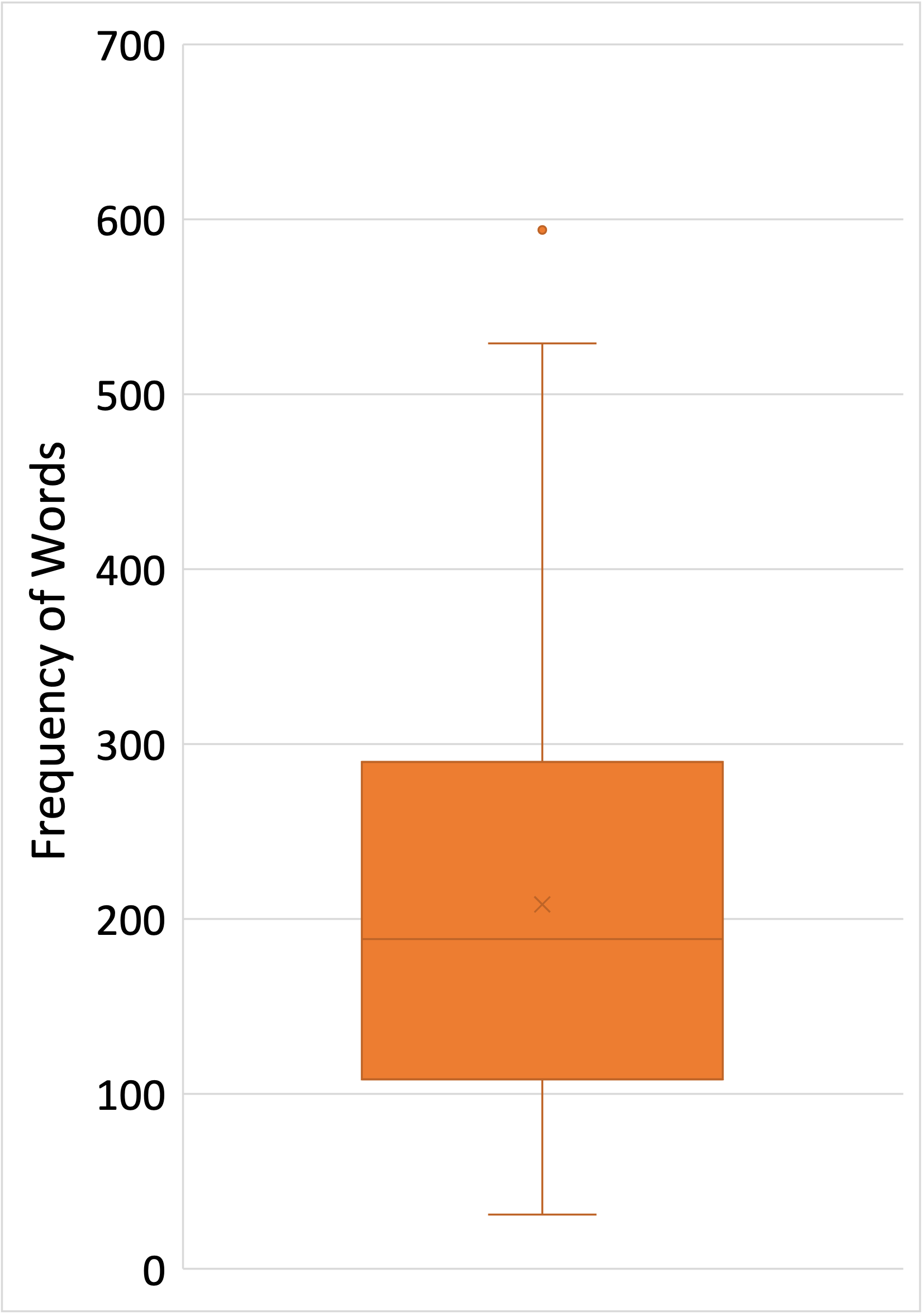}}}\hspace{5pt}
\subfloat[Pareto plot of \acrshort{procot} number of words]{%
\resizebox*{9cm}{!}{\includegraphics{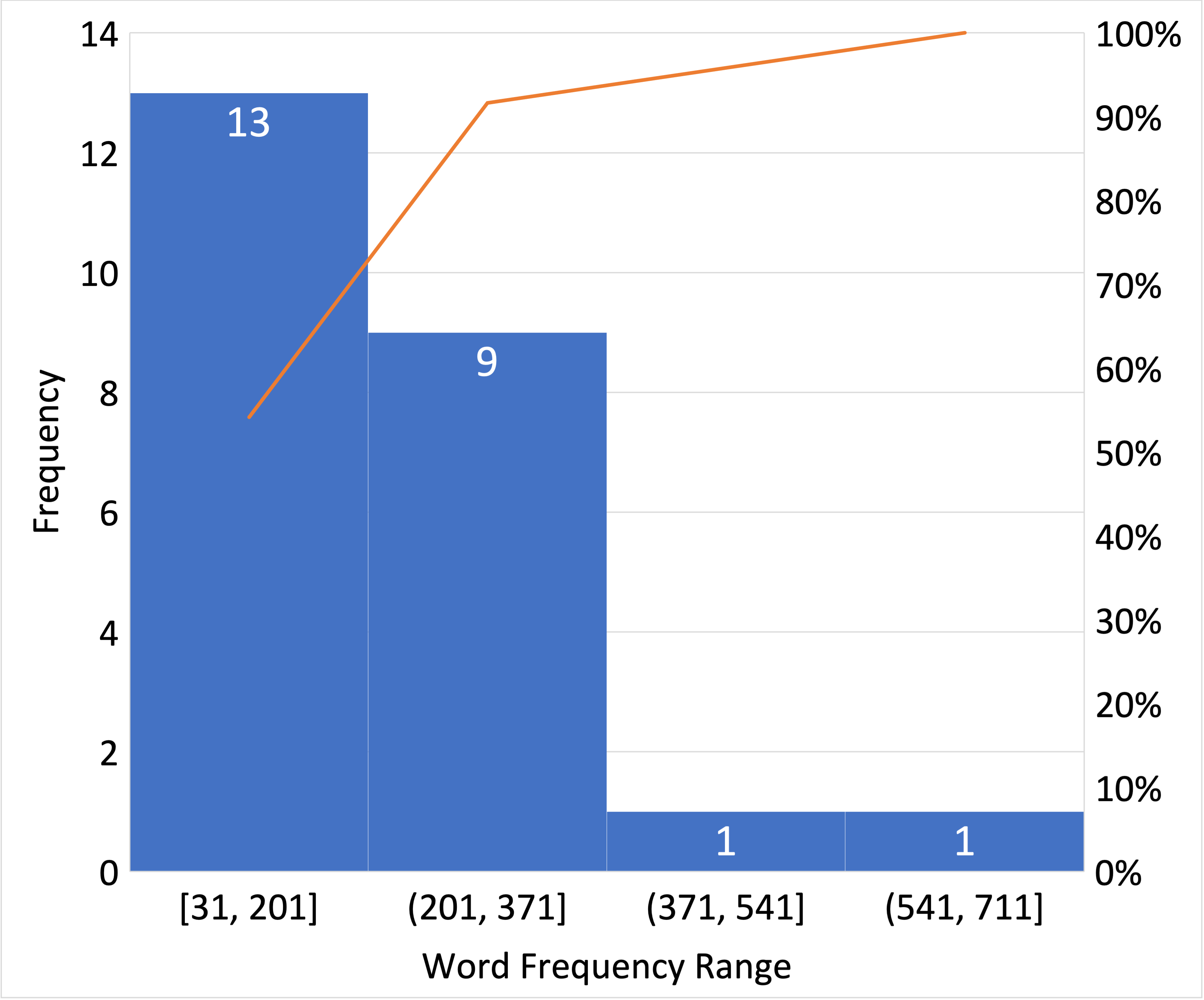}}}
\caption{Case 1: Quantitative plots of \acrshort{procot} number of words in student answers.} \label{plots1}
\end{figure}

The minimum, median, maximum, and average word counts for the students' \acrshort{procot} answers are 31, 189, 594, and 208, respectively, as shown in the Box plot of Figure \ref{plots1} (a).
Figure \ref{plots1} (b) shows the Pareto plot for the words in their answers, where most students used word counts in the range 31 and 201.
This implies most students prefer using fewer words to express their answers.
ChatGPT gave an average word count of 391 in its \acrshort{procot} answers while Phind gave 383.
These indicators of the average word counts by students and \acrshort{llm}s may be useful as cues for possible cheating, besides \acrshort{llm} artefacts that may result from copying an \acrshort{llm} output unedited.
This is because possible cheating through prompt engineering on an \acrshort{llm} to fix the word count is known to work poorly.\footnote{platform.openai.com/docs/guides/prompt-engineering/strategy-provide-reference-text} 
It is important to emphasize that word count alone is not enough to conclude that cheating has occurred.

\subsection{Case 2}
Out of the 39 submissions, 26 were valid because they followed the instructions, particularly with grounding their argument with peer-reviewed references.
In all the ChatGPT \acrshort{procot} answers\footnote{on 20 December, 2023}, only in-text citation is provided, when it does use references.
With regards to question 1 about \acrshort{llm} in the previous section, ChatGPT's \acrshort{procot} answer does not give any reference and merely regurgitates what was stated earlier rather than affirm or refute what was given.
With regards to question 2 about the vanishing gradient, ChatGPT's \acrshort{procot} answer gives 7 relevant references.
With regards to question 3 on transfer learning, ChatGPT's \acrshort{procot} answer gives 5 relevant references.
With regards to ChatGPT's answer to question 4 on similarity metrics, its \acrshort{procot} answer fails to follow the instruction, as it did not provide the minimum references, which is one of the reasons for rejecting some students' submissions.
BingAI, on the other hand, in all cases regurgitated the input and provided 4-6 blogs as references (when it provides references) instead of peer-reviewed articles.
It then adds a short paragraph as conclusion in 2 out of the 4 instances.
The appendix contains some of the examples.

\begin{figure}
\centering
\subfloat[Box and Whisker plot of \acrshort{procot} number of words]{%
\resizebox*{5cm}{!}{\includegraphics{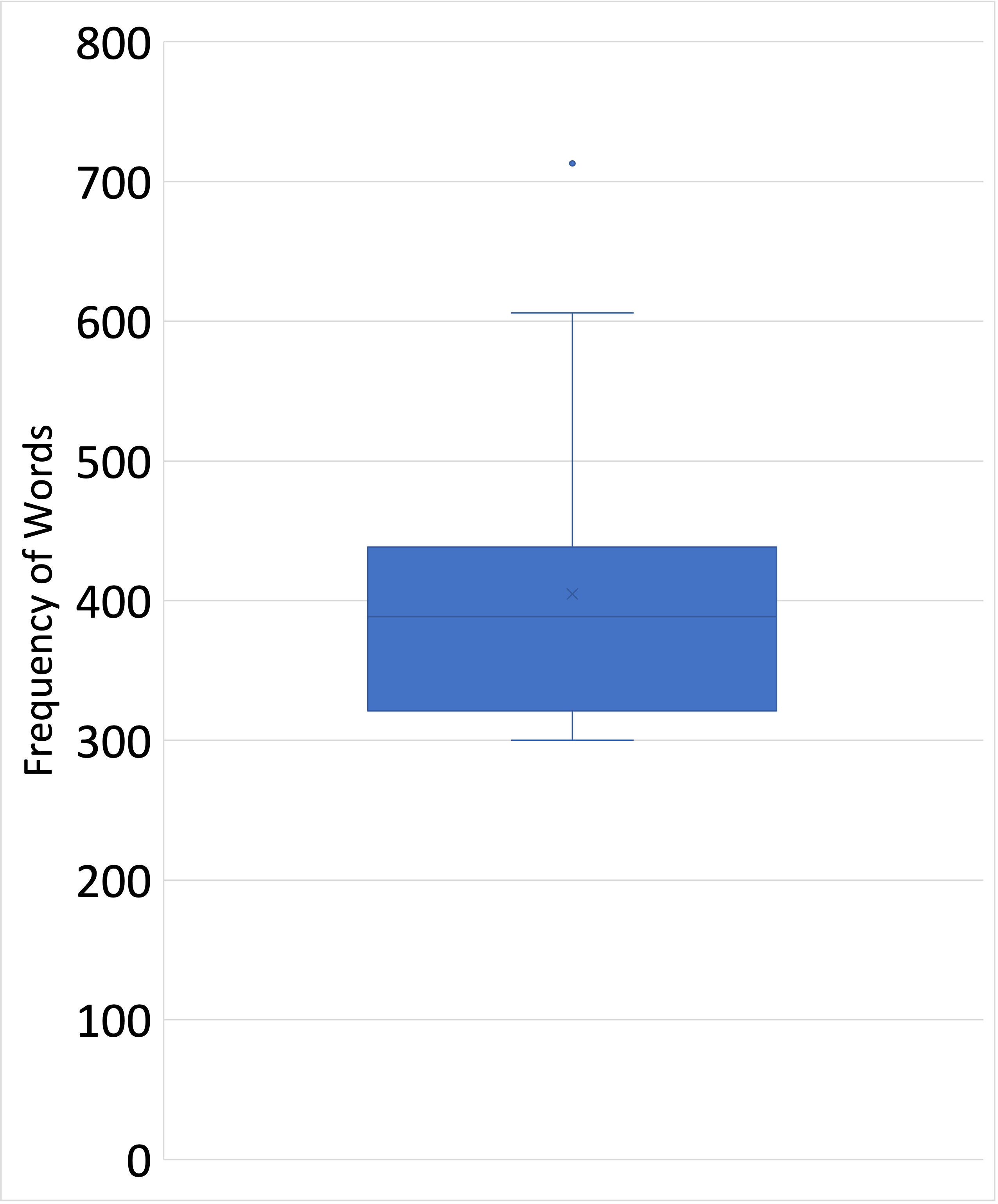}}}\hspace{5pt}
\subfloat[Pareto plot of \acrshort{procot} number of words]{%
\resizebox*{9cm}{!}{\includegraphics{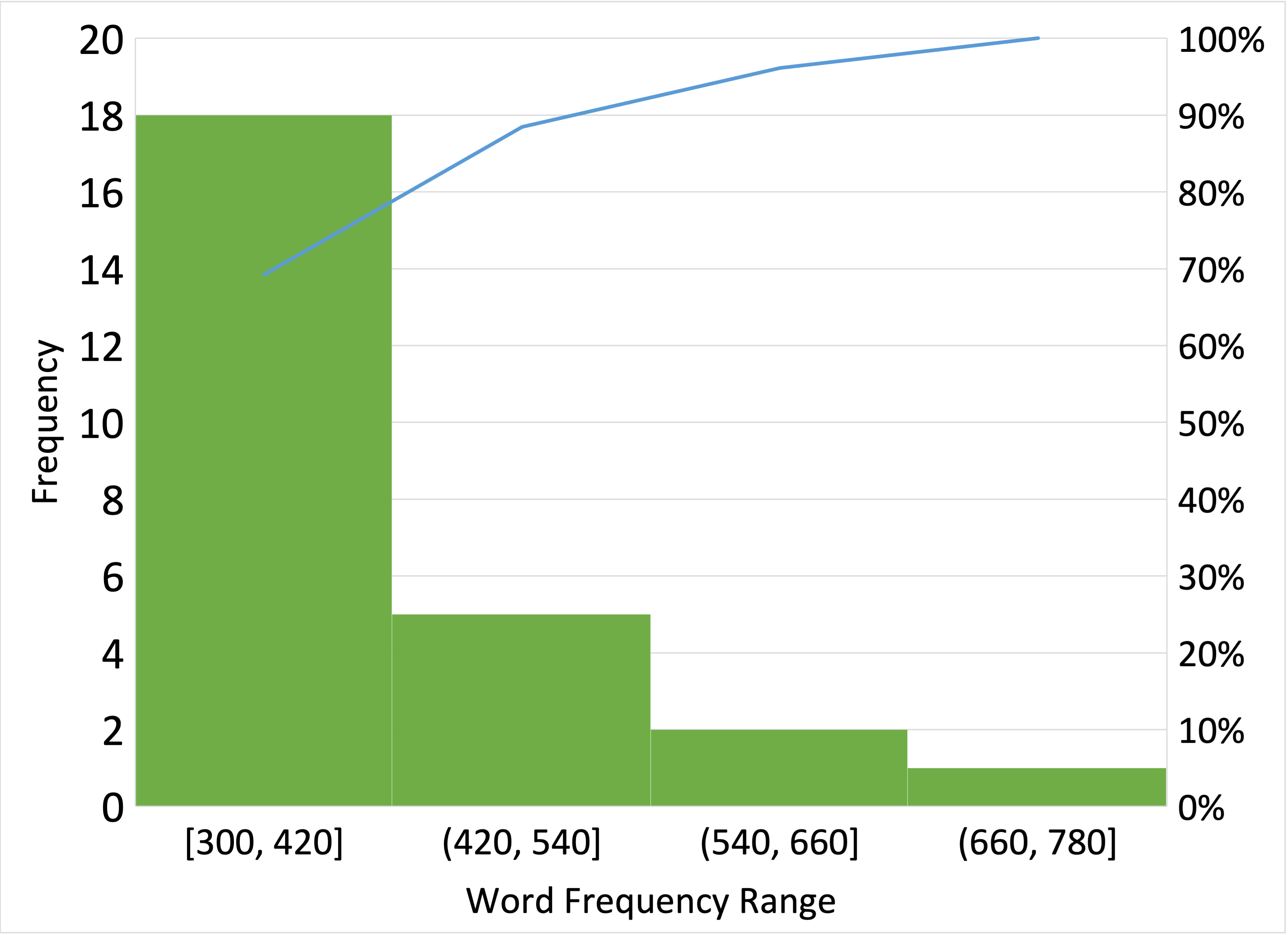}}}
\caption{Case 2: Quantitative plots of \acrshort{procot} number of words in student answers.} \label{plots2}
\end{figure}

Again, students use fewer words, as will be observed from Figure \ref{plots2}, preferring to align with the minimum word count of 300.
The median, maximum, and average word counts are 389, 713, and 405, respectively.
The average word counts for ChatGPT and BingAI are 356 and 315, respectively.
We assume the students' average word count is higher in this case because we required a minimum number of words.
It might be helpful to remove this restriction, as done in case 1, as a possible cue for cheating.

%\begin{table}
%\tbl{Example of a table showing that its caption is as wide as the table itself and justified.}
%{\begin{tabular}{lcccccc} \toprule
% & \multicolumn{2}{l}{Type} \\ \cmidrule{2-7}
% Class & One & Two & Three & Four & Five & Six \\ \midrule
% Alpha\textsuperscript{a} & A1 & A2 & A3 & A4 & A5 & A6 \\
% Beta & B2 & B2 & B3 & B4 & B5 & B6 \\
% Gamma & C2 & C2 & C3 & C4 & C5 & C6 \\ \bottomrule
%\end{tabular}}
%\tabnote{\textsuperscript{a}This footnote shows how to include
% footnotes to a table if required.}
%\label{sample-table}
%\end{table}

%For example, with coding, one way to evaluate \acrshort{llm}s is to check all the outputs from such \acrshort{llm} match what they are to give as output in a transparent manner. If the \acrshort{llm} under-delivers on the output then this negatively affects its transparency.
% Also, a student can be evaluated on the ability to adequately express a request (through prompt engineering) to an \acrshort{llm} and define the expected outputs, instead of coding.

\section{Conclusion}
\label{conc}
\acrshort{procot} shows that educators need not fear \acrshort{llm}s but embrace them as powerful tools to enrich the experiences of students in learning.
Weaknesses currently exist in the most popular free \acrshort{llm} - ChatGPT, and this makes it useful for teaching students to be better writers by grounding their writing in peer-reviewed references.
Observations from the results and ongoing research into \acrshort{llm}s imply these models will improve over time, especially as new data from methods like \acrshort{procot} can be used to retrain these models to be better.
They will become better grounded in their knowledge.
For example, we observe improvements in case 2 over case 1, which had one week difference between them, though it should be noted that the topics are different.

Hence, new methods may be required to evaluate students' writing performance and learning in the future.
We believe this is not a reason for educators to fear or bar students from using these tools but to see the opportunity to stimulate critical thinking through the engagement of students with these tools.
This will be possible through evolving, active and innovative methods.
Multimodal evaluation may be the solution in the future or methods that are grounded in interview/oral assessments.

\section*{Acknowledgement(s)}
This work is supported by the Wallenberg AI, Autonomous Systems and Software Program (WASP), funded by Knut and Alice Wallenberg Foundations and counterpart funding from Luleå University of Technology (LTU).

%\section*{Disclosure statement}

%An unnumbered section, e.g.\ \verb"\section*{Disclosure statement}", may be used to declare any potential conflict of interest and included \emph{in the non-anonymous version} before any Notes or References, after any Acknowledgements and before any Funding information.

%\section*{Funding}
%Part-funding for the writing of this publication was provided by the Computer Science, Electrical and Space Engineering Department's (SRT) Strategic Investmenet fund of Luleå University of Technology (LTU). 

\section*{Disclosure statement}
The authors report there are no competing interests to declare.

%An unnumbered section, e.g.\ \verb"\section*{Funding}", may be used for grant details, etc.\ if required and included \emph{in the non-anonymous version} before any Notes or References.

%\section*{Notes on contributor(s)}

%An unnumbered section, e.g.\ \verb"\section*{Notes on contributors}", may be included \emph{in the non-anonymous version} if required. A photograph may be added if requested.

%\section*{Nomenclature/Notation}

%An unnumbered section, e.g.\ \verb"\section*{Nomenclature}" (or \verb"\section*{Notation}"), may be included if required, before any Notes or References.

%\section*{Notes}

%An unnumbered `Notes' section may be included before the References (if using the \verb"endnotes" package, use the command \verb"\theendnotes" where the notes are to appear, instead of creating a \verb"\section*").

%\section{References}

%\begin{thebibliography}{}

%\end{thebibliography}
\bigskip

\bibliographystyle{apacite}
\bibliography{interactapasample}

\section*{Appendices}
\label{append}

%\processdelayedfloats %%% See above for an explanation of why this command might be needed here.

%\appendix 

\subsection*{Question 2 (Case 1):}
Who will benefit from AI in healthcare?

\paragraph*{ChatGPT Answer/Output:}
AI has the potential to benefit various stakeholders in the field of healthcare. Here are some of the key groups that stand to benefit from AI in health:

\begin{itemize}
    \item Patients:Improved Diagnostics: AI can assist in faster and more accurate disease diagnosis, leading to earlier and more effective treatments.
    \item Personalized Medicine: AI can help tailor treatment plans to individual patient profiles, increasing the likelihood of successful outcomes.
    \item Remote Monitoring: AI-powered devices and apps can enable remote monitoring of health conditions, allowing patients to better manage chronic illnesses and receive timely interventions.
    \item Enhanced Access: AI can improve access to healthcare services, especially in underserved or remote areas, through telemedicine and virtual health assistants.
    \item Healthcare Providers:Clinical Decision Support: AI can provide healthcare professionals with valuable insights and recommendations, aiding in diagnosis and treatment decisions.
    \item Administrative Efficiency: AI can streamline administrative tasks, such as appointment scheduling and medical coding, reducing the administrative burden on healthcare staff.
    \item Predictive Analytics: AI can predict disease outbreaks and patient needs, helping hospitals allocate resources more efficiently.
    \item Reduced Burnout: Automation of repetitive tasks can reduce the burnout experienced by healthcare workers.
    \item Researchers and Drug Development:Drug Discovery: AI accelerates drug discovery by identifying potential drug candidates and predicting their efficacy.
    \item Data Analysis: AI can analyze vast amounts of medical data to uncover patterns and insights that may lead to breakthroughs in medical research.
    \item Clinical Trials: AI can optimize patient recruitment for clinical trials and assist in monitoring trial data.
    \item Healthcare Payers and Insurance Companies:Improved Risk Assessment: AI can help insurers assess risk more accurately, leading to more personalized and competitive insurance plans.
    \item Fraud Detection: AI can detect and prevent healthcare fraud, saving costs for insurance companies.
    \item Population Health Management: AI can assist in managing the health of covered populations more effectively, reducing overall healthcare expenses.
    \item Governments and Public Health Organizations:Disease Surveillance: AI can help in monitoring disease outbreaks, facilitating rapid responses to public health emergencies.
\end{itemize}

Health Policy and Planning: AI can provide insights for policymakers to make data-driven decisions on healthcare infrastructure and resource allocation.
AI-driven home healthcare devices and monitoring systems can help caregivers and family members provide better care and peace of mind for their loved ones.
\\ Caregivers and Family Members:
While AI has the potential to provide numerous benefits in healthcare, it's important to address ethical, privacy, and security concerns to ensure that these technologies are used responsibly and for the benefit of all stakeholders in the healthcare ecosystem. Additionally, equitable access to AI-driven healthcare solutions and addressing potential biases are important considerations to ensure that the benefits of AI are distributed fairly.

\paragraph*{ChatGPT \acrshort{procot} Answer/Output:}
The statements are generally affirmed. Numerous studies and reviews in peer-reviewed literature support the potential benefits of AI in healthcare for various stakeholders. While I cannot provide direct references to specific articles, I can highlight the broad consensus from the literature.

Patients:
\begin{itemize}
    \item Improved Diagnostics: Studies have explored the use of AI in improving diagnostic accuracy across various medical domains. For example, AI models have been developed for early detection of diseases such as cancer and diabetic retinopathy.
    \item Personalized Medicine: AI-driven approaches, including machine learning algorithms, have been investigated for tailoring treatment plans based on individual patient characteristics. Personalized medicine is an active area of research in oncology and other fields.
    \item Remote Monitoring: AI-powered devices and applications have been studied for remote monitoring of health conditions. This includes wearable devices and smartphone apps that enable continuous monitoring of vital signs and chronic disease management.
    \item Enhanced Access: Telemedicine, powered by AI applications, has demonstrated potential in improving access to healthcare services, especially in underserved or remote areas.
\end{itemize}

Healthcare Providers:
\begin{itemize}
    \item Clinical Decision Support: The literature supports the use of AI for clinical decision support, providing insights and recommendations to healthcare professionals in diagnostics and treatment planning.
    \item Administrative Efficiency: Administrative tasks, such as appointment scheduling and medical coding, are areas where AI applications have shown promise in improving efficiency.
    \item Predictive Analytics: AI-driven predictive analytics has been explored for predicting disease outbreaks and optimizing resource allocation in healthcare settings.
    \item Reduced Burnout: Studies have investigated the impact of automation and AI in reducing the burden of repetitive tasks, contributing to the potential reduction in burnout among healthcare workers.
\end{itemize}

Researchers and Drug Development:
\begin{itemize}
    \item Drug Discovery: AI has been increasingly utilized in drug discovery, offering computational approaches for identifying potential drug candidates and predicting their efficacy.
    \item Data Analysis: The application of AI in analyzing large-scale medical data for research purposes has been well-documented in the literature.
    \item Clinical Trials: AI has been explored for optimizing patient recruitment in clinical trials and assisting in the monitoring of trial data.
\end{itemize}
Healthcare Payers and Insurance Companies:
\begin{itemize}
    \item Improved Risk Assessment: AI applications have been studied for more accurate risk assessment, contributing to the development of personalized insurance plans.
    \item Fraud Detection: AI has been employed for fraud detection in healthcare, aiding insurance companies in saving costs.
    \item Population Health Management: AI-driven approaches are investigated for population health management, aiming to enhance the effectiveness of healthcare services and reduce overall expenses.
\end{itemize}
Governments and Public Health Organizations:
\begin{itemize}
    \item Disease Surveillance: AI's role in disease surveillance and monitoring outbreaks has been acknowledged in the literature.
    \item Health Policy and Planning: The potential of AI to inform health policy and planning through data-driven insights has been discussed in the literature.
\end{itemize}

Caregivers and Family Members:
\begin{itemize}
    \item Home Healthcare Devices: AI-driven home healthcare devices and monitoring systems have been explored to provide better care and support for caregivers and family members.
\end{itemize}

Ethical Considerations: The statement emphasizes the importance of addressing ethical, privacy, and security concerns associated with AI in healthcare, which is a common consideration discussed in the literature. Additionally, equitable access and addressing biases are recognized as important aspects to ensure the fair distribution of AI benefits.

In conclusion, while specific references are not provided, the statements align with the broader consensus in the peer-reviewed literature on the potential benefits of AI in healthcare across various stakeholders. Researchers interested in specific studies on these topics may refer to reputable journals in healthcare informatics, medical imaging, bioinformatics, and related fields.

\paragraph*{Phind \acrshort{procot} Answer/Output:}

The statements provided are generally affirmative, as they align with the current understanding of how AI can benefit various stakeholders in the field of healthcare. However, it's important to note that while AI has the potential to provide numerous benefits, it's crucial to address ethical, privacy, and security concerns to ensure that these technologies are used responsibly and for the benefit of all stakeholders in the healthcare ecosystem.

\begin{enumerate}
    \item Patients: AI can indeed assist in faster and more accurate disease diagnosis, leading to earlier and more effective treatments. AI can also help in personalized medicine by tailoring treatment plans to individual patient profiles, increasing the likelihood of successful outcomes. AI-powered devices and apps can enable remote monitoring of health conditions, allowing patients to better manage chronic illnesses and receive timely interventions. AI can also improve access to healthcare services, especially in underserved or remote areas, through telemedicine and virtual health assistants.\textsuperscript{1}
    \item Healthcare Providers: AI can provide healthcare professionals with valuable insights and recommendations, aiding in diagnosis and treatment decisions. AI can streamline administrative tasks, such as appointment scheduling and medical coding, reducing the administrative burden on healthcare staff. AI can predict disease outbreaks and patient needs, helping hospitals allocate resources more efficiently. Automation of repetitive tasks can reduce the burnout experienced by healthcare workers.\textsuperscript{3}
    \item Researchers and Drug Development: AI accelerates drug discovery by identifying potential drug candidates and predicting their efficacy. AI can analyze vast amounts of medical data to uncover patterns and insights that may lead to breakthroughs in medical research. AI can optimize patient recruitment for clinical trials and assist in monitoring trial data.\textsuperscript{3}
    \item Healthcare Payers and Insurance Companies: AI can help insurers assess risk more accurately, leading to more personalized and competitive insurance plans. AI can detect and prevent healthcare fraud, saving costs for insurance companies. AI can assist in managing the health of covered populations more effectively, reducing overall healthcare expenses.\textsuperscript{3}
    \item Governments and Public Health Organizations: AI can help in monitoring disease outbreaks, facilitating rapid responses to public health emergencies. AI can provide insights for policymakers to make data-driven decisions on healthcare infrastructure and resource allocation. AI-driven home healthcare devices and monitoring systems can help caregivers and family members provide better care and peace of mind for their loved ones.\textsuperscript{3}
    \item Caregivers and Family Members: AI can provide numerous benefits in healthcare, but it's important to address ethical, privacy, and security concerns to ensure that these technologies are used responsibly and for the benefit of all stakeholders in the healthcare ecosystem. Additionally, equitable access to AI-driven healthcare solutions and addressing potential biases are important considerations to ensure that the benefits of AI are distributed fairly.
\end{enumerate}

\subsection*{Question 4 (Case 2):}
Describe the two paradigms of using a pre-trained model in transfer learning: feature extraction and fine-tuning.

\paragraph*{BingAI Answer/Output:}

Certainly! In the context of transfer learning, there are two primary paradigms for utilizing a pre-trained model: 

Feature Extraction: In this approach, we leverage the representations learned by a previously trained network to extract meaningful features from new data. 
The process involves adding a new classifier (usually a fully connected layer) on top of the pre-trained model. 
The pre-trained model’s convolutional base (which contains generic feature maps) remains frozen during this step. 
By doing so, we can repurpose the feature maps learned previously for a different dataset without retraining the entire model. 
However, the final classification layer of the pre-trained model is specific to its original task and class labels. 
Example: Suppose we use a pre-trained model trained on a large-scale image classification task and add a new classifier for a specific set of classes (e.g., cats vs. dogs). 

Fine-Tuning: Fine-tuning involves unfreezing a few of the top layers of the pre-trained model’s base. 
We then jointly train both the newly-added classifier layers and the last layers of the base model. 
The goal is to adapt the higher-order feature representations in the base model to make them more relevant for the specific task. 
Fine-tuning is particularly useful when we have sufficient labeled examples for the target task. 
Example: After feature extraction, we fine-tune the base model’s top layers to improve its performance on the specific classification task. 	

In summary, feature extraction allows us to reuse learned features, while fine-tuning refines the model for a specific task by adjusting its higher-level representations.\textsuperscript{1} 

\paragraph*{ChatGPT \acrshort{procot} Answer/Output:}

The provided information accurately describes two common paradigms in transfer learning: feature extraction and fine-tuning. These approaches are widely employed to leverage pre-trained models for new tasks. The assertions made are supported by multiple references from peer-reviewed articles and books.

Feature extraction involves using a pre-trained model's convolutional base to extract meaningful features from new data. This process is well-documented in the literature. For instance, in the book "Deep Learning" by Ian Goodfellow et al. (2016), the authors discuss the concept of transfer learning and feature extraction in the context of convolutional neural networks (CNNs).
The idea of freezing the convolutional base during feature extraction is consistent with the work of Razavian et al. (2014) in the paper "CNN Features Off-the-shelf: an Astounding Baseline for Recognition." The study explores using pre-trained CNNs for image classification tasks by extracting features from the convolutional layers.
The example provided, using a pre-trained model for a large-scale image classification task and adding a new classifier for specific classes (e.g., cats vs. dogs), aligns with the practical application of feature extraction, as discussed by Yosinski et al. (2014) in "How transferable are features in deep neural networks?"

Fine-tuning, as described, involves unfreezing a few of the top layers of the pre-trained model and jointly training both the new classifier layers and the last layers of the base model. This process is extensively discussed in the literature. For instance, in the paper "Visualizing and Understanding Convolutional Networks" by Zeiler and Fergus (2014), the authors explore fine-tuning strategies to improve the performance of pre-trained CNNs on specific tasks.
The concept that fine-tuning is particularly useful when there are sufficient labeled examples for the target task is supported by the work of Donahue et al. (2014) in "Decaf: A Deep Convolutional Activation Feature for Generic Visual Recognition." The study demonstrates the effectiveness of fine-tuning with ample labeled examples for image classification.

In summary, the information provided accurately describes the feature extraction and fine-tuning paradigms in transfer learning, and the statements align with findings from various peer-reviewed sources. These paradigms offer practical strategies for utilizing pre-trained models and adapting them to specific tasks.

\paragraph*{BingAI \acrshort{procot} Answer/Output:}

In the context of transfer learning, there are two primary paradigms for utilizing a pre-trained model:

Feature Extraction and Fine-Tuning.
Feature Extraction involves leveraging the representations learned by a previously trained network to extract meaningful features from new data. The process involves adding a new classifier (usually a fully connected layer) on top of the pre-trained model. The pre-trained model’s convolutional base (which contains generic feature maps) remains frozen during this step. By doing so, we can repurpose the feature maps learned previously for a different dataset without retraining the entire model. However, the final classification layer of the pre-trained model is specific to its original task and class labels. For example, suppose we use a pre-trained model trained on a large-scale image classification task and add a new classifier for a specific set of classes (e.g., cats vs. dogs).\textsuperscript{123}

Fine-Tuning involves unfreezing a few of the top layers of the pre-trained model’s base. We then jointly train both the newly-added classifier layers and the last layers of the base model. The goal is to adapt the higher-order feature representations in the base model to make them more relevant for the specific task. Fine-tuning is particularly useful when we have sufficient labeled examples for the target task. For example, after feature extraction, we fine-tune the base model’s top layers to improve its performance on the specific classification task.\textsuperscript{123}

In summary, feature extraction allows us to reuse learned features, while fine-tuning refines the model for a specific task by adjusting its higher-level representations.\textsuperscript{123}

%\section{Troubleshooting}

%\section{Obtaining the template and class file}

%\subsection{Via the Taylor \& Francis website}

%\subsection{Via e-mail}

%This article template, the \texttt{interact} class file and the associated open-source \LaTeX\ packages are also available via e-mail. Requests should be addressed to \texttt{latex.helpdesk@tandf.co.uk}, clearly stating for which journal you require the template and class file.

\end{document}